\documentclass[runningheads]{llncs}
\PassOptionsToPackage{table}{xcolor}

\usepackage{eccv}

\usepackage{eccvabbrv}
\usepackage{multirow}
\usepackage{microtype}
\usepackage{graphicx}
\usepackage{enumitem}
\usepackage{subcaption}
\usepackage{booktabs}
\usepackage{tabularx}
\usepackage{amsmath,amssymb,mathtools}
\usepackage{placeins}   
\usepackage{dblfloatfix} 

\usepackage[accsupp]{axessibility}  

\usepackage{hyperref}

\usepackage{orcidlink}

\begin{document}

\title{Global Logic and Local Search: Dual-Stream Multimodal In-Context Learning for Verifiable Industrial Anomaly Detection} 

\titlerunning{GLLS for Verifiable Industrial Anomaly Detection}

\author{Runzhi Deng\orcidlink{0009-0000-6119-8549}\inst{1}$^*$ \and
Yundi Hu\orcidlink{0009-0004-9936-3828}\inst{1}$^*$ \and
Yiming Zhong\orcidlink{0009-0003-5355-0460}\inst{1}$^*$ \and
Zhao Wang\orcidlink{0009-0008-5831-7496}\inst{2}$^\dagger$ \and
Xixi Liu\orcidlink{0000-0002-3032-524X}\inst{2} \and
Hongsong Wang\orcidlink{0000-0002-9464-1778}\inst{3} \and
Caifeng Shan\orcidlink{0000-0002-2131-1671}\inst{1} \and
Fang Zhao\orcidlink{0000-0002-6772-8042}\inst{1}$^\dagger$}

\authorrunning{R. Deng et al.}

\institute{
Nanjing University, Nanjing, Jiangsu, China \and
China Mobile Zijin Innovation Institute, Nanjing, Jiangsu, China \and
Southeast University, Nanjing, Jiangsu, China
}

\maketitle
\begingroup
\let\thefootnote\relax
\footnotetext{\textsuperscript{*} Equal contribution.}
\footnotetext{\textsuperscript{\textdagger} Corresponding authors.}
\endgroup
\setcounter{footnote}{0}

\begin{figure}[tb]
  \centering
  \includegraphics[width=\linewidth]{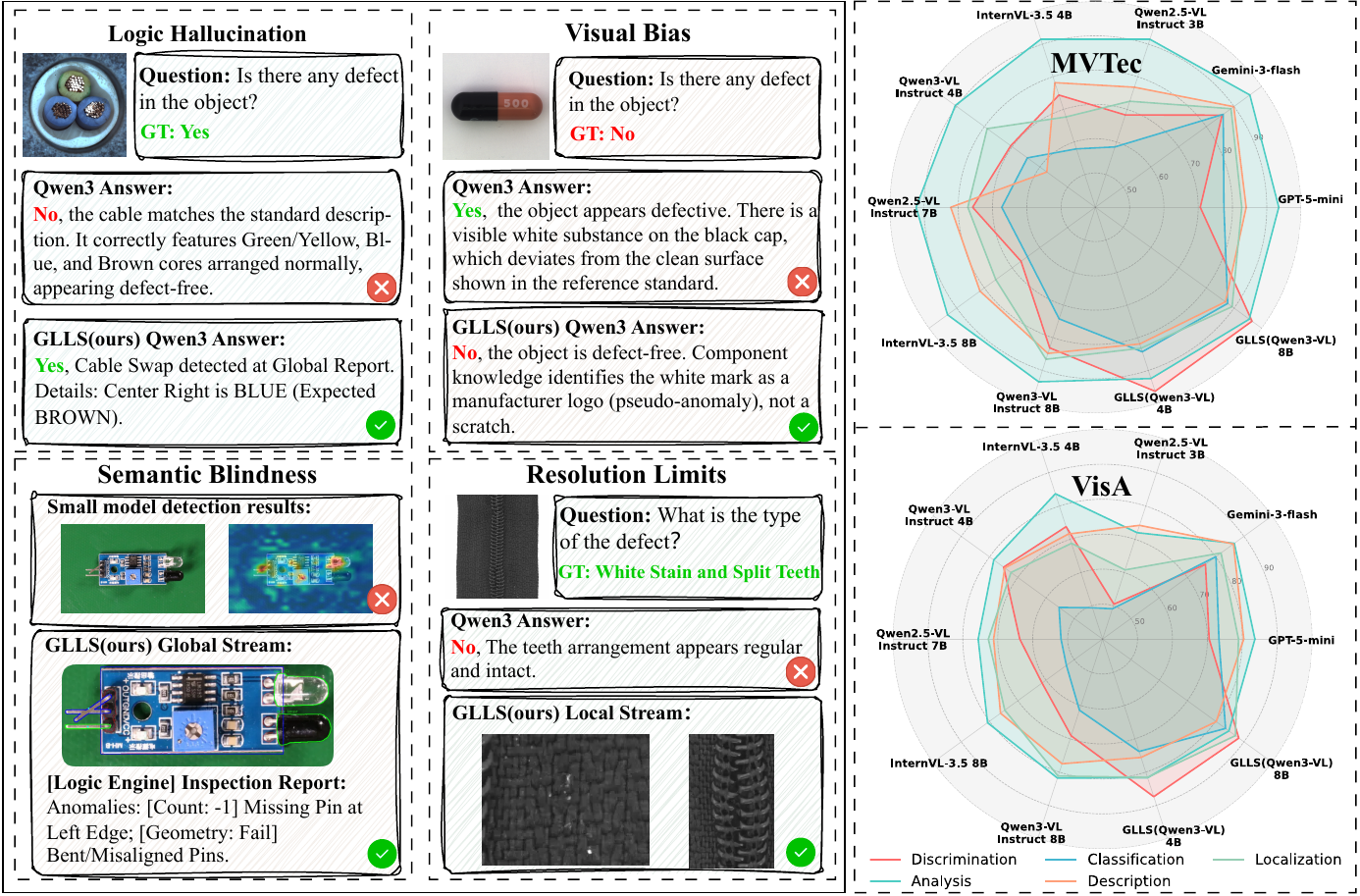}
  \caption{\textbf{Motivation and Performance Overview.} 
  The left panels visualize critical failure modes in industrial inspection, where LMMs typically suffer from \textbf{Logic Hallucination} and \textbf{Visual Bias} (top), while specialized models of smaller scale struggle with \textbf{Semantic Blindness} and \textbf{Resolution Limits} (bottom). We address these limitations by incorporating a tool-grounded \textit{Global Logic} stream and an active \textit{Local Search} stream to support auditable inspection. The right panels present quantitative analyses showing that the GLLS framework improves over representative general-purpose baselines under a controlled budget of computation across five metrics on the MMAD-QA benchmark.}
  \label{fig:motivation}
\end{figure}
\begin{abstract}
Large Multimodal Models (LMMs) show strong few-shot generalization, but industrial anomaly detection remains difficult because defects are small, input resolution is limited, and textual standards are not always grounded in visual evidence. Recent optimization-based methods improve alignment through fine-tuning, but they often require many defective samples, which are unavailable in early deployment. We present \textbf{Global Logic and Local Search (GLLS)}, a training-free framework for reference-guided multimodal in-context verification. GLLS uses a Part-Aware Visual-Logical Atlas to organize normal references and structured specifications in the inference context. It combines a \textbf{Global \& Logic Stream}, where SAM 3 extracts partially checkable visual facts, with a \textbf{Fine-Grained \& Actions Stream}, where MCTS selects local evidence crops under a fixed budget. Experiments on MMAD-QA and additional anomaly detection datasets show consistent gains over matched and general-purpose baselines, while keeping the final diagnostic decision traceable to explicit visual evidence throughout the inspection trace.
\end{abstract}

\section{Introduction}
\label{sec:intro}

Industrial anomaly detection fundamentally differs from general recognition tasks because it functions as a rigorous compliance audit against explicit specifications rather than open-ended inference. Unlike open-world scenarios where models must infer normality from large-scale data, industrial inspection requires the identification of deviations from rigorous standards, such as the topology of components and geometric tolerances. This challenge is particularly acute during the initial phase of deployment, where defective samples are non-existent, normal references are scarce, and the system must rely heavily on natural language specifications. While Large Multimodal Models (LMMs) exhibit remarkable generalization capabilities \cite{hurst2024gpt,bai2025qwen3}, the direct application of these models to this task remains challenging. They inherently struggle to ground abstract textual standards into verification at the pixel level and frequently fail to resolve fine-grained defects due to inherent limitations in input resolution.

To bridge the gap between abstract standards and visual inspection, recent studies have explored various adaptation strategies. Recent optimization-based methods attempt to align LMMs via reinforcement learning, multi-agent frameworks, or extensive supervised fine-tuning to enforce consistency in reasoning~\cite{li2025iad, chao2025anomalyr1, miao2025agentiad, ji2025autoiad, guan2025emit, zhao2025omniad}. However, these approaches are computationally intensive and demand extensive data, necessitating numerous defective samples that are typically unavailable during the initial phases of deployment. To mitigate this data dependency, logical reasoning and retrieval-enhanced methods such as Echo \cite{chen2025can} and Triad \cite{li2025triad} utilize external bases of knowledge or expert-guided tokenizers \cite{wei2022chain, jin2025logicad, li2025lad}. Nevertheless, these methods typically rely on static context injection \cite{yu2024visrag} or generic region proposals, which fail to synthesize the global logic of assembly and the local sensory evidence required for auditable inspection.

In contrast, human experts manage this complexity through a cognitive process described by the \textit{Reverse Hierarchy Theory} of vision \cite{hochstein2002view}.  Inspectors do not strictly rely on massive training data; instead, they consult technical specifications to form a high-level understanding of structures, and subsequently direct their attention to verify specific details via active perception \cite{bajcsy2018revisiting}, echoing recent MLLM evidence that explicit perception tokens enhance visual reasoning \cite{bigverdi2025perception}. This interplay between explicit domain knowledge and active visual search allows humans to audit complex assemblies even with minimal prior exposure to visual samples. We argue that a reliable industrial inspection system can benefit from an active and knowledge-grounded workflow rather than relying on passive, end-to-end mechanisms of perception.

Toward this end, we propose a training-free framework, termed \textbf{Global Logic and Local Search (GLLS)}, which is motivated by two practical requirements in cold-start industrial inspection. First, effective inspection requires decoupling tool-grounded structural checks from probabilistic visual reasoning to ensure that logical constraints are explicitly evaluated and remain fully auditable. Second, resolving ambiguity in high-resolution images benefits from an evidence-seeking strategy that selects informative observations under a fixed budget rather than relying only on passive retrieval. GLLS treats the LMM as an auditor orchestrated by a \textbf{Part-Aware Visual-Logical Atlas (PVLA)}. This atlas leverages natural language specifications and available reference samples to construct an executable plan for inspection, providing explicit chains of evidence for the final diagnostic conclusions.

The contributions of this work are summarized as follows:
\begin{itemize}
    \item We propose GLLS, a training-free dual-stream framework that orchestrates LMMs with auditable intermediate evidence by synergizing a \textbf{Global \& Logic Stream} for tool-grounded structural checks and a \textbf{Fine-Grained \& Actions Stream} for the acquisition of fine-grained sensory evidence.
    \item We construct the Part-Aware Visual-Logical Atlas (PVLA) to organize natural-language specifications and normal references into an explicit graph of inspection knowledge. It supports text-guided zero-shot inspection, where SAM 3 \cite{carion2025sam} grounds logical rules without visual references, and reference-guided few-shot verification when normal samples are available.
    \item We design a budgeted active Local Search mechanism utilizing Monte Carlo Tree Search (MCTS) \cite{browne2012survey},  which selects spatially distinct evidence crops under fixed computation and context budgets without exhaustive scanning.
    \item The GLLS framework achieves strong performance on the MMAD-QA benchmark and demonstrates strong generalization across various industrial datasets, including MPDD, DTD, and DAGM. By strictly controlling budgets of information and computation, our framework provides a competitive, training-free alternative to optimization-based methods, exhibiting consistent efficacy across data regimes ranging from text-guided zero-shot settings to reference-guided few-shot settings.
\end{itemize}
\section{Related Work}
\label{sec:related}

\noindent\textbf{Generalized Multimodal Anomaly Detection.}
Current research on aligning Multimodal Models with industrial standards encompasses both optimization-based strategies and lightweight adaptations. To enforce consistent reasoning chains, reinforcement learning approaches, such as IAD-R1~\cite{li2025iad}, AnomalyR1~\cite{chao2025anomalyr1}, and EMIT~\cite{guan2025emit}, utilize Group Relative Policy Optimization to align outputs with human logic, while LR-IAD~\cite{zeng2025lr} employs specialized reward functions to address class imbalance. In parallel with these RL-based methods, architectural modifications characterize frameworks such as EIAD~\cite{zhang2025eiad} and Triad~\cite{li2025triad}, which decouple dialog generation or incorporate expert tokenizers to focus on critical regions. Furthermore, extensive fine-tuning frameworks, including OmniAD~\cite{zhao2025omniad} and AgentIAD~\cite{miao2025agentiad}, have been proposed to enhance anomaly understanding. On the spectrum of data efficiency, smaller adaptations, such as AdaptCLIP~\cite{gao2025adaptclip} and ABounD~\cite{deng2025abound}, leverage CLIP-based architectures for boundary-driven learning, but they often lack the deep semantic reasoning capabilities required for complex diagnosis. Despite these contributions, these methods generally necessitate either extensive defective training data or costly gradient updates. Consequently, this dependency restricts their applicability during initial deployment phases, where only a few normal reference images and natural language specifications are available.

\noindent\textbf{Multimodal In-Context Learning.}
In-Context Learning (ICL) enables models to adapt to new tasks via demonstrations without parameter updates~\cite{huang2024multimodal}. However, standard Multimodal ICL often exhibits strong biases toward textual priors and may neglect fine-grained visual details when modalities contradict~\cite{baldassini2024makes}. Recent research emphasizes compositional learning to improve context understanding~\cite{li2024improving} and to mitigate these hallucinations. However, generic retrieval-augmented generation and domain injection techniques~\cite{yu2024visrag, song2025injectingdomainspecificknowledgelarge, mecklenburg2024injecting} risk introducing noisy or irrelevant context, which degrades precision. We address these gaps through a reference-guided multimodal ICL framework, wherein a Part-Aware Visual-Logical Atlas injects precise executable standards into the inference context. This approach ensures that the reasoning process is grounded in auditable canonical norms rather than hallucinatory priors, thereby enabling immediate deployment without parameter updates.

\section{Method}
\label{sec:method}
\begin{figure*}[t]
  \centering
  \includegraphics[width=\linewidth]{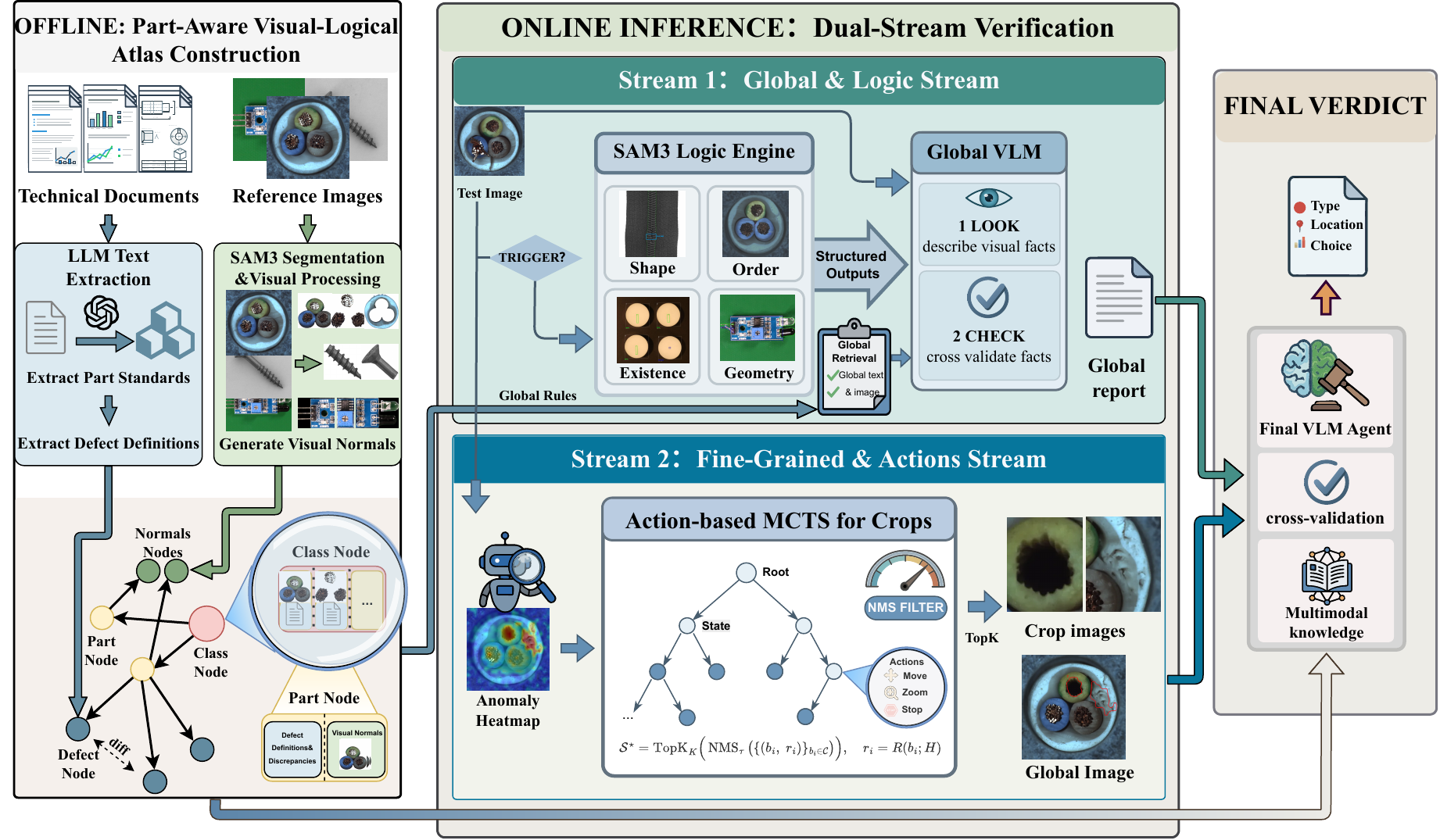}
  \caption{\textbf{System Overview.} 
  \textbf{Offline Construction:} The \textbf{Part-Aware Visual-Logical Atlas (PVLA)} is constructed from natural language specifications and a few normal reference images. The SAM 3 segmentation model extracts visual primitives to populate the atlas with canonical visual norms and executable logic.
  \textbf{Online Dual-Stream Inference:} 
  (1) The \textbf{Global \& Logic Stream} executes tool-grounded checks via the SAM 3 model to validate structural integrity.
  (2) In parallel, the \textbf{Fine-Grained \& Actions Stream} formulates the inspection process as a budgeted Markov Decision Process (MDP), utilizing \textbf{MCTS} to actively zoom in on ambiguous regions.
  (3) A Vision-Language Model (VLM) performs the final \textbf{Reference-Guided In-Context Verification} by fusing structural verdicts and local evidence chains.}
  \label{fig:pipeline}
\end{figure*}
\subsection{Overview of the GLLS Framework}
\label{sec:overview}

Industrial anomaly detection demands a rigorous standard of reliability that purely data-driven foundation models often fail to meet, because a fluent description offers no guarantee that critical constraints, such as component presence or assembly order, are correctly checked. To bridge the gap between probabilistic description and auditable inspection, we propose \textbf{Global Logic and Local Search (GLLS)}, which decouples evidence acquisition from verdict reasoning. The framework uses a pre-compiled rulebook to guide the LMM as an evidence-driven auditor, rather than passively consuming images.

The workflow consists of two distinct phases. In the offline phase, we employ a hybrid pipeline to construct a \textbf{Part-Aware Visual-Logical Atlas (PVLA)} by synthesizing structured dataset priors with natural language reasoning into an executable knowledge graph. This creates a grounded memory bank where every physical component is linked to specific inspection rules. Subsequently, in the online phase, the system executes a dual-stream inference strategy. The \textbf{Global \& Logic Stream} utilizes SAM 3~\cite{carion2025sam} to perform tool-grounded validation of structural logical constraints, while the \textbf{Fine-Grained \& Actions Stream} uses Monte Carlo Tree Search (MCTS) as a budgeted proposal mechanism for selecting high-resolution evidence regions. By converging these streams, the final verdict is a reasoned decision grounded in both tool-grounded checks and fine-grained visual evidence.

\subsection{Hybrid Atlas Construction}
\label{sec:construction}

To ensure the framework remains scalable while maintaining strict alignment with industrial standards, we use a hybrid construction pipeline that expands sparse dataset labels into structured inspection specifications. This construction process integrates structured domain priors with the reasoning capabilities of Large Multimodal Models and the open-vocabulary segmentation power of state-of-the-art vision models.

\paragraph{Knowledge Expansion via GPT-5}
We utilize the structured metadata provided by the MMAD benchmark as the initial logical skeleton, which provides the ground-truth set of normal component categories specific to the target domain. Since these raw labels lack the physical granularity required for detailed inspection, we employ \textbf{GPT-5}~\cite{openai2025gpt5} to expand the seed taxonomy into candidate physical definitions based on the provided natural language ontology. Specifically, the model proposes granular sub-component attributes and potential failure modes derived from the component definitions. To ensure reliability, the generated definitions are filtered to mitigate inconsistencies. For instance, given a \textit{Cable} component, GPT-5 infers the existence of sub-parts such as the \textit{Outer Sheath} and \textit{Copper Cores}, while simultaneously hypothesizing probable anomalies like \textit{Insulation Cuts} or \textit{Wire Deformation} based on textual context. This process helps reduce the semantic gap between abstract category names and concrete physical constraints.

\paragraph{Visual Grounding with SAM 3}
To populate the visual nodes of the atlas with canonical reference features, we leverage the SAM 3 segmentation model \cite{carion2025sam}, which provides strong text-conditioned grounding for short part phrases and produces candidate masks. We use the sub-component names generated by GPT-5 as textual prompts to guide SAM 3 in extracting regions of interest from normal reference images. To improve the reliability of these visual primitives, we filter the extracted segments using fixed confidence, IoU, and mask-stability criteria. This filtering step reduces ambiguous or low-quality proposals before they are inserted into the atlas.

\paragraph{Semantic Graph Formalism}
Formally, we define the resulting Part-Aware Visual-Logical Atlas as a heterogeneous directed graph $\mathcal{G} = (\mathcal{V}, \mathcal{E})$ constructed to explicitly decouple canonical standards from anomalous deviations. To capture the hierarchical assembly and state duality, we formulate the vertex set $\mathcal{V}$ as the disjoint union of four semantic subsets as
\begin{equation}
    \mathcal{V} = \{v_{cls}\} \cup \mathcal{V}_{part} \cup \mathcal{V}_{norm} \cup \mathcal{V}_{def}
\end{equation}
At the root level, $v_{cls}$ represents the abstract object category, which decomposes into a set of physical components $\mathcal{V}_{part} = \{v_{p_i}\}_{i=1}^{N}$ that mirror the topological structure of the target object. To model the ideal state, we define the normal reference set $\mathcal{V}_{norm} = \{v_{n_i}\}_{i=1}^{N}$, where each $v_{n_i}$ denotes the normal-state anchor associated with component $v_{p_i}$. Each anchor stores the textual standard and, when available, one or more reference visual features extracted by SAM 3. When multiple normal references are available, they are stored as separate alternatives rather than averaged, preserving intra-class variation during retrieval. Deviations are captured by the defect set $\mathcal{V}_{def} = \{v_{d_j}\}_{j=1}^{M}$, which enumerates candidate failure modes inferred by GPT-5.

The relational logic is governed by the edge set $\mathcal{E}$, which is partitioned into three functional mappings as
\begin{equation}
    \mathcal{E} = \mathcal{E}_{comp} \cup \mathcal{E}_{state} \cup \mathcal{E}_{diff}
\end{equation}
First, the \textit{hierarchical decomposition edges} $\mathcal{E}_{comp}$ connect the class node $v_{cls}$ to constituent parts in $\mathcal{V}_{part}$ to establish the physical assembly. Second, the \textit{state association edges} $\mathcal{E}_{state}$ branch from each part node $v_{p_i}$ to establish a dual connection comprising a bijective mapping to the unique normal reference $v_{n_i}$ for standardization, and a one-to-many mapping to relevant defect nodes in $\mathcal{V}_{def}$ for anomaly enumeration. Finally, the \textit{differentiation edges} $\mathcal{E}_{diff} \subset \mathcal{V}_{def} \times \mathcal{V}_{def}$ explicitly connect visually confusing defect patterns. These edges are dynamically generated by GPT-5 during the offline knowledge expansion phase to encode contrastive rules (e.g., differentiating a ``scratch'' from a ``crack'' via spatial depth cues), thereby enforcing fine-grained discriminative boundaries and mitigating ambiguity during online verification.

\subsection{Dual-Stream Verifiable Inference}
\label{sec:inference}

To reconcile the conflicting requirements of global structural consistency and fine-grained defect sensitivity, we propose a dual-stream architecture where two parallel pathways maximize inspection throughput.

\subsubsection{Global \& Logic Stream}
\label{sec:stream_1}

The Global \& Logic Stream focuses on detecting macroscopic structural anomalies and logical inconsistencies by performing a preliminary audit of the entire object. This process uses a SAM 3-based fact extractor to transform industrial specifications into partially checkable visual primitives, which effectively grounds the inspection in measurable physical evidence. The global verification is formulated as a structured consistency check defined by
\begin{equation}
\begin{aligned}
    \mathcal{F}_{\text{fact}} &= \Psi_{\text{SAM}}(I, \mathcal{R}_{\text{logic}}) \\
    v_{\text{struct}} &= \mathcal{V}_{\text{audit}}(I, \mathcal{F}_{\text{fact}}, \mathcal{K}_{\text{glob}} \mid \mathcal{P}_{\text{chain}})
\end{aligned}
\end{equation}
where $I$ represents the query image and $\mathcal{R}_{\text{logic}}$ denotes the set of categorical rules used to guide fact extraction. $\mathcal{R}_{\text{logic}}$ is instantiated from PVLA part rules, where predicates are paired with fixed SAM text phrases. $\Psi_{\text{SAM}}$ masks the named parts, and the resulting visual rules produce $\mathcal{F}_{\text{fact}}$, including component counts, spatial arrangement, color order, and coarse geometry. The term $\mathcal{K}_{\text{glob}}$ denotes the multimodal standards, including a background-free global reference image and textual specifications; in zero-shot settings, it reduces to textual specifications only. The structural verdict $v_{\text{struct}}$ is an intermediate report rather than the final diagnosis. It is derived by a multimodal auditor $\mathcal{V}_{\text{audit}}$ operating under a sequential prompting policy $\mathcal{P}_{\text{chain}}$, which first asks the model to identify visual facts from the target image $I$ independently of the standard. These findings are then cross-validated against the logic report $\mathcal{F}_{\text{fact}}$ and the multimodal global knowledge $\mathcal{K}_{\text{glob}}$ to produce the intermediate structural report. By transforming abstract checks into a text-grounded fact-matching task, this stream provides a more grounded and partially checkable basis for the subsequent fine-grained analysis.

\subsubsection{Fine-Grained \& Actions Stream}
\label{sec:stream_2}

Exhaustive high-resolution inspection for minute defects is computationally prohibitive. Thus, we formulate fine-grained inspection as a budgeted Markov Decision Process (MDP), where MCTS is used as a budgeted proposal mechanism for selecting evidence crops \cite{browne2012survey, jaech2024openai, guo2025deepseek, wang2025pixel, zheng2025deepeyes}.

The state $s_t$ corresponds to an active image crop parameterized by its bounding box. The action space $\mathcal{A}$ includes discrete translations, scaling, and an inspect region action triggered by global proposals. To guide the search under a fixed budget, we use an anomaly heatmap $\mathcal{H}$ from a lightweight localizer as a heuristic prior. For each state, we compute a heatmap-derived search score $S_{\text{search}}$, which summarizes the local saliency of the crop and its alignment with global proposals. Rather than serving as a new search algorithm, MCTS is used here to maintain exploration over neighboring views and retain spatially diverse evidence crops under the same budget. It performs selection, expansion/simulation, and backpropagation; after exhausting the simulation budget, we apply NMS to obtain final proposals:

\textbf{1. Selection:} Starting from the root, the policy selects child nodes maximizing a modified Upper Confidence Bound (UCB) that balances exploitation, exploration, and the heuristic prior:
\begin{equation}
a_t = \arg\max_{a \in \mathcal{A}} \left( \frac{V(s_{next})}{N(s_{next})} + C_{p} \sqrt{\frac{2 \ln N(s_t)}{N(s_{next})}} + \lambda S_{\text{search}}(s_{next}) \right)
\end{equation}
where $V(\cdot)$ and $N(\cdot)$ are the accumulated value and visit count, $C_p$ is the exploration constant, and $\lambda = 0.5$ weights the heatmap-derived search prior $S_{\text{search}}$ to bias the search towards visually deviant areas.

\textbf{2. Expansion \& Simulation:} Valid actions expand non-terminal nodes. The new leaf node $s_{leaf}$ is evaluated via a thresholded reward derived from the heatmap:
\begin{equation}
R(s_{leaf}) = \begin{cases} S_{\text{search}}(s_{leaf}) & \text{if } S_{\text{search}}(s_{leaf}) \geq \tau_{sim} \\ 0 & \text{otherwise} \end{cases}
\end{equation}
where $\tau_{sim} = 0.05$ is a permissive low-confidence gate for pruning uninformative regions rather than a final anomaly threshold; it is fixed across all experiments.

\textbf{3. Backpropagation:} The scalar reward $R(s_{leaf})$ is propagated upward through the traversed path to update the node statistics. For each node $s$ in the path, the values are updated as $V(s) \leftarrow V(s) + R(s_{leaf})$ and $N(s) \leftarrow N(s) + 1$.

\textbf{4. NMS-based Proposal Filtering:} After the budget is exhausted, we apply NMS over visited node coordinates (IoU threshold 0.1) using the MCTS posterior score to rank candidates. The top-$K$ spatially distinct crops form the final evidence set $\mathcal{I}_{crops}$. Compared with directly selecting the maximum or Top-$K$ heatmap peaks, this procedure favors spatially distinct views and yields modest but consistent gains under the same crop budget.

\subsubsection{Reference-Guided In-Context Verification}
\label{sec:verdict}

In the final stage, the outputs from the parallel streams converge to produce a verified decision. The VLM acts as the final verifier by combining the structural report $v_{\text{struct}}$ and the fine-grained evidence crops $\mathcal{I}_{crops}$ with the structured context retrieved from the \textbf{Part-Aware Visual-Logical Atlas (PVLA)}. Unlike generic retrieval baselines that rely on flat similarity searches, our mechanism exploits the \textbf{hierarchical graph structure} of the PVLA to retrieve component-scoped standards and reference evidence.

The final verdict $y$ is generated by maximizing the consistency between the observed evidence and the retrieved standards via
\begin{equation}
    y^* = \arg\max_{y \in \mathcal{Y}} P(y \mid v_{\text{struct}}, \mathcal{I}_{crops}, \mathcal{Z}_{graph}, \mathcal{T}_{graph})
\end{equation}
In MMAD-QA, $\mathcal{Y}$ is the closed set of answer options for the current question. The verifier is prompted to output the option in the required format, e.g., ``The correct answer is (X)'', and the parser extracts $X$ as the final prediction. Thus, the argmax denotes closed-set VLM selection rather than an additional learned optimization module. Here, $P(\cdot)$ formalizes the conditional log-likelihood of the Large Multimodal Model generating the correct diagnostic class sequence $y$, parameterized by the fused prompt. Crucially, the context retrieval constitutes a topological traversal rather than a simple flat vector search. The system navigates from the root object node down to specific component nodes (e.g., \textit{Cable} $\to$ \textit{Sheath}) to fetch the precise executable context $\mathcal{T}_{graph}$ and canonical visual normals $\mathcal{Z}_{graph}$ relevant to the current view.

In few-shot settings, this hierarchical alignment allows the model to perform localized comparison against spatially corresponding normal reference samples. Conversely, in zero-shot settings where $\mathcal{Z}_{graph} = \emptyset$, the verification relies on the structured text logic encoded in the leaf nodes. These nodes encode discriminative attributes including the \textit{Visual Signature} and \textit{Contrast Logic} to differentiate between defect types. This allows the VLM to ground evidence crops $\mathcal{I}_{crops}$ into semantic definitions without requiring visual references, while keeping the final output auditable through an evidence chain that points to the visual evidence and failure mode supporting the decision.
\section{Experiments}
\label{sec:exp}

\subsection{Setup}
We evaluate GLLS on the \textbf{MMAD-QA} benchmark~\cite{jiang2024mmad}, specifically selecting the MVTec-AD~\cite{bergmann2019mvtec} and VisA~\cite{zou2022spot} subsets. To verify cross-domain robustness, we also include MPDD~\cite{jezek2021deep}, DTD~\cite{aota2023zero}, and DAGM~\cite{wieler2007weakly} datasets. Our evaluation focuses on five core metrics relevant to industrial inspection: Anomaly Discrimination, Defect Analysis, Classification, Description, and Localization. Since MMAD-QA is formulated as closed-set visual question answering, many tasks require choosing among fixed answer options rather than producing calibrated anomaly scores. We therefore report task accuracy for the five QA metrics. For cross-dataset binary anomaly-detection experiments, we follow the binary classification accuracy protocol used by the compared methods. Threshold-independent metrics such as AUROC and PRAUC are not uniformly defined for all closed-set QA tasks and are therefore not used as primary metrics. For matched ablations, we fix the VLM backbone, support policy, evidence budget, and verifier so that the effect of PVLA retrieval, dual-stream verification, and local search can be isolated.
\paragraph{Implementation Details}
\label{sec:implementation}

We implement GLLS using the PyTorch framework on 8$\times$RTX4090 (48GB) GPUs. For the VLM backbones, we utilize the official instruction-tuned checkpoints of Qwen2.5-VL (7B) and Qwen3-VL (4B, 8B) without any parameter updates. The \textbf{Global \& Logic Stream} integrates the SAM3 model to execute deterministic segmentation checks. To support the Fine-Grained Action Stream, we employ AdaptCLIP~\cite{gao2025adaptclip} in zero-shot settings and ABounD~\cite{deng2025abound} in one-shot settings as the lightweight localizer to generate the anomaly heatmap $\mathcal{H}$ used for heuristic guidance. In the Fine-Grained \& Actions Stream, the MCTS is configured with a simulation count of $N=50$ per decision step and an exploration constant $C_{p}=2.0$, with the maximum evidence budget fixed at $K_{max}=3$ crops based on our ablation analysis. We employ standard chat templates for all VLM interactions. All external modules are frozen and used only as evidence providers or heuristic priors. GPT-5 is used only during offline PVLA construction, SAM 3 provides text-conditioned segmentation for visual facts, and AdaptCLIP/ABounD provide heatmap priors for local search. The final diagnostic answer is produced by the VLM verifier from the retrieved PVLA context, structural report, and evidence crops.
\begin{table}[t]
\centering
\newcommand{\inc}[1]{\textsuperscript{\textcolor{teal}{+#1}}}
\newcommand{\dec}[1]{\textsuperscript{\textcolor{red}{-#1}}}

\caption{Evaluation on the \textbf{MMAD benchmark}~\cite{jiang2024mmad} across MVTec and VisA domains. Results reflect the accuracy on specific MMAD-QA tasks. Gains and losses relative to the base model are indicated as superscripts (\textcolor{teal}{green} for increase, \textcolor{red}{red} for decrease). The best results are highlighted in bold.}
\label{tab:main}

\scriptsize 
\renewcommand{\arraystretch}{1.1} 

\resizebox{\textwidth}{!}{%
\begin{tabular}{l@{\hspace{4pt}}c@{\hspace{4pt}}cccccc@{\hspace{6pt}}cccccc}
\toprule
\multirow{2}{*}{\textbf{Model}} & \multirow{2}{*}{\textbf{Param}} & \multicolumn{6}{c}{\textbf{MVTec}} & \multicolumn{6}{c}{\textbf{VisA}} \\
\cmidrule(lr){3-8} \cmidrule(lr){9-14}
 &  & \textbf{Disc.} & \textbf{Ana.} & \textbf{Cls.} & \textbf{Desc.} & \textbf{Loc.} & \textbf{Avg.} & \textbf{Disc.} & \textbf{Ana.} & \textbf{Cls.} & \textbf{Desc.} & \textbf{Loc.} & \textbf{Avg.} \\
\midrule
GPT-5-nano & / & 73.5 & 91.8 & 63.8 & 75.7 & 73.9 & 75.7 & 70.1 & 78.0 & 55.2 & 72.5 & 68.8 & 68.9 \\
GPT-5-mini & / & 70.6 & 93.5 & 77.5 & 84.0 & 82.7 & 81.7 & 70.5 & 83.6 & 73.3 & 80.4 & 78.9 & 77.3 \\
Gemini-2.5 & / & 71.8 & 90.3 & 72.4 & 78.7 & 80.6 & 78.8 & 67.1 & 80.3 & 68.2 & 78.7 & 72.7 & 73.4 \\
Gemini-3 & / & 85.7 & \textbf{95.7} & 86.1 & \textbf{89.9} & 88.9 & 89.3 & 76.4 & \textbf{86.6} & 80.1 & \textbf{86.3} & 81.9 & 82.3 \\
\midrule
Qwen2.5-VL & 3B & 68.3 & 91.5 & 58.5 & 76.8 & 72.5 & 73.5 & 50.5 & 72.0 & 49.1 & 74.2 & 60.8 & 61.3 \\
InternVL-3.5 & 4B & 74.4 & 91.5 & 57.9 & 78.2 & 67.7 & 73.9 & 73.8 & 83.7 & 49.4 & 71.6 & 68.8 & 69.5 \\
MiniCPM-V4 & 4B & 79.4 & 94.5 & 68.0 & 87.2 & 86.2 & 83.1 & 68.9 & 77.7 & 51.0 & 75.3 & 77.1 & 70.0 \\
Qwen3-VL & 4B & 70.5 & 90.5 & 64.5 & 57.5 & 79.0 & 72.4 & 75.0 & 78.5 & 55.4 & 74.6 & 72.4 & 71.2 \\
Qwen2.5-VL & 7B & 75.8 & 92.3 & 67.3 & 82.2 & 77.2 & 79.0 & 63.7 & 75.6 & 51.9 & 71.2 & 72.7 & 67.0 \\
InternVL-3.5 & 8B & 66.8 & 93.3 & 64.5 & 81.6 & 75.9 & 76.4 & 61.4 & 80.7 & 52.8 & 76.1 & 74.6 & 69.1 \\
Qwen3-VL & 8B & 83.1 & 93.5 & 74.2 & 84.8 & 86.6 & 84.4 & 69.1 & 81.8 & 61.4 & 77.5 & 81.2 & 74.2 \\
Qwen3-VL-Thinking & 8B & 81.7 & 93.7 & 69.7 & 83.5 & 81.7 & 82.1 & 66.4 & 78.8 & 59.3 & 79.6 & 76.2 & 72.1 \\
MiniCPM-V4.5 & 9B & 77.8 & 93.5 & 72.5 & 89.4 & 80.3 & 82.7 & 61.0 & 81.9 & 61.1 & 79.0 & 70.1 & 70.6 \\
\midrule

\rowcolor{gray!10} GLLS (Qwen3-VL, \textit{0-shot}) & 4B & 82.6\inc{12.1} & 91.3\inc{0.8} & 79.8\inc{15.3} & 77.3\inc{19.8} & 77.1\dec{-1.9} & 81.6\inc{9.2} & 75.4\inc{0.4} & 78.2\dec{-0.3} & 71.5\inc{16.1} & 78.7\inc{4.1} & 78.5\inc{6.1} & 76.5\inc{5.3} \\

\rowcolor{gray!10} GLLS (Qwen2.5-VL, \textit{0-shot}) & 7B & 84.0\inc{8.2} & 92.4\inc{0.1} & 73.3\inc{6.0} & 83.6\inc{1.4} & 77.4\inc{0.2} & 82.1\inc{3.1} & 72.6\inc{8.9} & 75.7\inc{0.1} & 62.2\inc{10.3} & 71.0\dec{-0.2} & 71.6\dec{-1.1} & 70.6\inc{3.6} \\

\rowcolor{gray!10} GLLS (Qwen3-VL, \textit{0-shot}) & 8B & 88.6\inc{5.5} & 93.9\inc{0.4} & 81.9\inc{7.7} & 85.0\inc{0.2} & 86.2\dec{-0.4} & 87.1\inc{2.7} & 74.3\inc{5.2} & 80.9\dec{-0.9} & 65.2\inc{3.8} & 75.5\dec{-2.0} & 79.8\dec{-1.4} & 75.1\inc{0.9} \\

\rowcolor{gray!10} GLLS (Qwen3-VL, \textit{1-shot}) & 4B & 96.4\inc{25.9} & 92.5\inc{2.0} & 84.3\inc{19.8} & 81.9\inc{24.4} & 83.2\inc{4.2} & 87.7\inc{15.3} & 87.4\inc{12.4} & 81.7\inc{3.2} & 73.8\inc{18.4} & 75.6\inc{1.0} & 81.6\inc{9.2} & 80.0\inc{8.8} \\

\rowcolor{gray!10} GLLS (Qwen2.5-VL, \textit{1-shot}) & 7B & 93.1\inc{17.3} & 92.7\inc{0.4} & 78.2\inc{10.9} & 83.5\inc{1.3} & 79.8\inc{2.6} & 85.5\inc{6.5} & 82.9\inc{19.2} & 81.6\inc{6.0} & 68.4\inc{16.5} & 76.2\inc{5.0} & 80.7\inc{8.0} & 78.0\inc{11.0} \\

\rowcolor{gray!10} GLLS (Qwen3-VL, \textit{1-shot}) & 8B & \textbf{96.5}\inc{13.4} & 95.5\inc{2.0} & \textbf{87.7}\inc{13.5} & 86.9\inc{2.1} & \textbf{89.3}\inc{2.7} & \textbf{91.2}\inc{6.8} & \textbf{88.2}\inc{19.1} & 84.7\inc{2.9} & \textbf{83.5}\inc{22.1} & 80.2\inc{2.7} & \textbf{87.0}\inc{5.8} & \textbf{84.7}\inc{10.5} \\
\bottomrule
\end{tabular}%
}
\end{table}
\begin{table}[t]
\centering
\caption{Cross-dataset generalization performance in industrial anomaly detection excluding VisA. We report binary classification accuracy (\%). * denotes threshold optimized using Youden's J statistic. \textbf{GLLS (Ours)} variants are listed in the final section.}
\label{tab:generalization_glls}

\scriptsize 
\renewcommand{\arraystretch}{1.0} 

\begin{tabular}{llccccc}
\toprule
 & \textbf{Model} & \textbf{Scale} & \textbf{MPDD} & \textbf{DTD} & \textbf{DAGM} & \textbf{Avg.} \\
\midrule
\multirow{6}{*}{\textbf{CLIP-based}} 
 & WinClip\cite{jeong2023winclip} ($\tau = 0.5$) & - & 72.10 & 25.90 & 86.90 & 57.08 \\
 & WinClip\cite{jeong2023winclip}* & - & 41.80 & 90.00 & 88.90 & 75.85 \\
 & AnomalyClip\cite{zhou2024anomalyclip} ($\tau = 0.5$) & - & 56.49 & 81.71 & 22.19 & 54.24 \\
 & AnomalyClip\cite{zhou2024anomalyclip}* & - & 68.23 & 88.91 & 95.36 & 83.67 \\
 & AdaptCLIP\cite{gao2025adaptclip}($\tau = 0.5$) & - & 63.98 & 84.00 & 82.17 & 76.72 \\
 & AdaptCLIP\cite{gao2025adaptclip}* & - & 72.16 & 91.89 & 94.37 & 86.14 \\
\midrule
\multirow{3}{*}{\textbf{Proprietary}} 
 & GPT4o-mini & - & 67.90 & 90.56 & 90.80 & 80.81 \\
 & GPT4o & - & 69.43 & 90.80 & 90.17 & 81.04 \\
 & GPT-4.1 & - & 66.81 & 90.18 & 89.93 & 81.98 \\
\midrule
\multirow{5}{*}{\textbf{Open-source}} 
 & LLaVA-NeXT & 34B & 61.57 & 72.62 & 23.15 & 53.70 \\
 & InternVL2 & 76B & 66.38 & 90.58 & 94.52 & 80.32 \\
 & InternVL3 & 8B & 70.74 & 85.05 & 81.33 & 75.54 \\
 & InternVL3 & 38B & 69.43 & 90.79 & 89.87 & 79.71 \\
 & Qwen2.5-VL & 7B & 68.34 & 85.59 & 94.83 & 76.07 \\
\midrule
\multirow{2}{*}{\textbf{Fine-tuned}} & IAD-R1\cite{li2025iad} (LLaVA-OV) & 7B & 70.90 & 96.20 & 94.80 & 86.10 \\
& AD-FM\cite{liao2025ad}(Qwen2.5-VL) & 7B & 72.71 & 92.64 & 95.51 & 84.57 \\
\midrule

\rowcolor{gray!10}
& GLLS (Qwen3-VL, \textit{0-shot}) & 4B & 72.34 & 93.12 & 96.32 & 87.26 \\
\rowcolor{gray!10}
& GLLS (Qwen2.5-VL, \textit{0-shot}) & 7B & 71.56 & 88.34 & 95.43 & 85.11 \\
\rowcolor{gray!10}
& GLLS (Qwen3-VL, \textit{0-shot}) & 8B & 75.41 & 93.40 & 97.15 & 88.65 \\
\rowcolor{gray!10}
& GLLS (Qwen3-VL, \textit{1-shot}) & 4B & 78.45 & 93.56 & 96.65 & 89.55 \\
\rowcolor{gray!10}
& GLLS (Qwen2.5-VL, \textit{1-shot}) & 7B & 73.10 & 91.45 & 96.12 & 86.89 \\
\rowcolor{gray!10}
\multirow{-6}{*}{\textbf{Ours}} & GLLS (Qwen3-VL, \textit{1-shot}) & 8B & \textbf{80.23} & \textbf{95.31} & \textbf{97.48} & \textbf{91.01} \\
\bottomrule
\end{tabular}
\end{table}
\begin{table}[tb]
\centering
\caption{Ablation study of the proposed GLLS method on MVTec and VisA datasets. We compare the Base model, Offline-only (Knowledge Retrieval), Online-only (Two-stream Verification), and the full GLLS method. The checkmark (\checkmark) indicates the inclusion of the corresponding module. Best results for each model are bolded.}
\resizebox{\textwidth}{!}{%
\begin{tabular}{lccccccccccccccc}
\toprule
\multirow{2}{*}{\textbf{Model}} & \multirow{2}{*}{\textbf{Param}} & \multicolumn{2}{c}{\textbf{Strategy}} & \multicolumn{6}{c}{\textbf{MVTec}} & \multicolumn{6}{c}{\textbf{VisA}} \\
\cmidrule(lr){3-4} \cmidrule(lr){5-10} \cmidrule(lr){11-16}
 &  & \textbf{Offline} & \textbf{Online} & \textbf{Disc.} & \textbf{Ana.} & \textbf{Cls.} & \textbf{Desc.} & \textbf{Loc.} & \textbf{Avg.} & \textbf{Disc.} & \textbf{Ana.} & \textbf{Cls.} & \textbf{Desc.} & \textbf{Loc.} & \textbf{Avg.} \\
\midrule
\multirow{4}{*}{Qwen3-VL-Instruct} & \multirow{4}{*}{4B} &  &  & 70.5 & 90.5 & 64.5 & 57.5 & 79.0 & 72.40 & 75.0 & 78.5 & 55.4 & 74.6 & 72.4 & 71.18 \\
 &  & \checkmark &  & 88.7 & 92.4 & 84.1 & 81.8 & 81.6 & 85.72 & 74.7 & 81.5 & 72.3 & \textbf{77.6} & 76.9 & 76.60 \\
 &  &  & \checkmark & 95.2 & 90.6 & 74.4 & 79.7 & 82.1 & 84.40 & 87.1 & 78.9 & 56.6 & 74.8 & 79.7 & 75.42 \\
 &  & \checkmark & \checkmark & \textbf{96.4} & \textbf{92.5} & \textbf{84.3} & \textbf{81.9} & \textbf{83.2} & \textbf{87.66} & \textbf{87.4} & \textbf{81.7} & \textbf{73.8} & 75.6 & \textbf{81.6} & \textbf{80.02} \\
\midrule
\multirow{4}{*}{Qwen2.5-VL-Instruct} & \multirow{4}{*}{7B} &  &  & 75.8 & 92.3 & 67.3 & 82.2 & 77.2 & 78.96 & 63.7 & 75.6 & 51.9 & 71.2 & 72.7 & 67.02 \\
 &  & \checkmark &  & 80.2 & 86.8 & 74.9 & 74.4 & 68.9 & 77.04 & 64.0 & 68.0 & 57.9 & 64.0 & 60.9 & 62.96 \\
 &  &  & \checkmark & 91.6 & 90.2 & 69.4 & 76.5 & \textbf{81.6} & 81.86 & \textbf{84.5} & 76.8 & 56.6 & 70.4 & 80.4 & 73.74 \\
 &  & \checkmark & \checkmark & \textbf{93.1} & \textbf{92.7} & \textbf{78.2} & \textbf{83.5} & 79.8 & \textbf{85.46} & 82.9 & \textbf{79.6} & \textbf{68.4} & \textbf{76.2} & \textbf{80.7} & \textbf{77.56} \\
\midrule
\multirow{4}{*}{Qwen3-VL-Instruct} & \multirow{4}{*}{8B} &  &  & 83.1 & 93.5 & 74.2 & 84.8 & 86.6 & 84.44 & 69.1 & 81.8 & 61.4 & 77.5 & 81.2 & 74.20 \\
 &  & \checkmark &  & 89.7 & 94.4 & 87.5 & 85.8 & 87.5 & 88.98 & 77.7 & 81.5 & 78.9 & 79.5 & 81.7 & 79.86 \\
 &  &  & \checkmark & 93.8 & 93.1 & 75.4 & 82.6 & 85.8 & 86.14 & 87.5 & 82.0 & 69.2 & 77.5 & 83.6 & 79.96 \\
 &  & \checkmark & \checkmark & \textbf{96.5} & \textbf{95.5} & \textbf{87.7} & \textbf{86.9} & \textbf{89.3} & \textbf{91.18} & \textbf{88.2} & \textbf{84.7} & \textbf{83.5} & \textbf{80.2} & \textbf{87.0} & \textbf{84.72} \\
\bottomrule
\end{tabular}%
}

\label{tab:ablation}
\end{table}
\begin{table}[!t]
    \centering
    \caption{Ablation Studies on Core Architectural Components. (a) Step-wise ablation demonstrating the contribution of the Online Dual-Stream architecture. (b) Ablation on the Part-Aware Visual-Logical Atlas (PVLA) isolating the contribution of the Offline Construction strategy.}
    \label{tab:core_ablations}
    
    \scriptsize
    
    \begin{subtable}[t]{0.48\linewidth}
        \centering
        \caption{Impact of Online Dual-Stream}
        \label{tab:stream_ablation}
        \renewcommand{\arraystretch}{1.32}
        \begin{tabular}{llccc}
            \toprule
            \textbf{Model} & \textbf{Config.} & \textbf{MVTec} & \textbf{VisA} & \textbf{Avg.} \\
            \midrule
            \multirow{3}{*}{\shortstack{Qwen3\\4B}} 
             & Base (Offline) & 85.7 & 76.6 & 81.2 \\
             & + Fine-Grained & 86.8 & 78.5 & 82.7 \\
             & \textbf{GLLS} & \textbf{87.7} & \textbf{80.0} & \textbf{83.8} \\
            \midrule
            \multirow{3}{*}{\shortstack{Qwen2.5\\7B}} 
             & Base (Offline) & 77.0 & 63.0 & 70.0 \\
             & + Fine-Grained & 82.3 & 73.1 & 77.7 \\
             & \textbf{GLLS} & \textbf{85.5} & \textbf{77.6} & \textbf{81.5} \\
            \midrule
            \multirow{3}{*}{\shortstack{Qwen3\\8B}} 
             & Base (Offline) & 89.0 & 79.9 & 84.4 \\
             & + Fine-Grained & 89.9 & 82.6 & 86.2 \\
             & \textbf{GLLS} & \textbf{91.2} & \textbf{84.7} & \textbf{88.0} \\
            \bottomrule
        \end{tabular}
    \end{subtable}
    \hfill
    \begin{subtable}[t]{0.48\linewidth}
        \centering
        \caption{Impact of Offline PVLA}
        \label{tab:pvla_ablation}
        \renewcommand{\arraystretch}{1.0}
        \begin{tabular}{llccc}
            \toprule
            \textbf{Model} & \textbf{Method} & \textbf{MVTec} & \textbf{VisA} & \textbf{Avg.} \\
            \midrule
            \multirow{4}{*}{\shortstack{Qwen3\\4B}} 
             & Base & 72.4 & 71.2 & 71.8 \\
             & Text-Guided & 80.5 & 73.2 & 76.9 \\
             & Image-Guided & 82.1 & 74.1 & 78.1 \\
             & \textbf{PVLA} & \textbf{85.7} & \textbf{76.6} & \textbf{81.2} \\
            \midrule
            \multirow{4}{*}{\shortstack{Qwen2.5\\7B}} 
             & Base & \textbf{79.0} & \textbf{67.0} & \textbf{73.0} \\
             & Text-Guided & 77.8 & 65.5 & 71.7 \\
             & Image-Guided & 76.5 & 64.1 & 70.3 \\
             & \textbf{PVLA} & 77.0 & 63.0 & 70.0 \\
            \midrule
            \multirow{4}{*}{\shortstack{Qwen3\\8B}} 
             & Base & 84.4 & 74.2 & 79.3 \\
             & Text-Guided & 86.5 & 76.4 & 81.5 \\
             & Image-Guided & 87.2 & 78.1 & 82.7 \\
             & \textbf{PVLA} & \textbf{89.0} & \textbf{79.9} & \textbf{84.4} \\
            \bottomrule
        \end{tabular}
    \end{subtable}
\end{table}
\subsection{Main Results}
Table~\ref{tab:main} compares our full system against a diverse set of representative general-purpose Multimodal Large Language Models, including the Qwen-VL, InternVL and MiniCPM-V~\cite{bai2025qwen3, bai2025qwen2, wang2025internvl3, yao2024minicpm, yu2026minicpm}. GLLS generally improves over the corresponding baselines across both MVTec-AD and VisA subsets. Notably, the gains are not limited to a single stage: the full pipeline improves discrimination reliability while also strengthening defect type classification and region localization. This suggests that the Global \& Logic Stream and the Fine-Grained \& Actions Stream provide complementary evidence for adapting general VLMs to industrial inspection. Table~\ref{tab:main} evaluates general-purpose VLM baselines under the same MMAD-QA protocol, while the ablation tables provide matched comparisons under the same backbone and evidence budget. While we do not explicitly compare against optimization-based methods that have not been benchmarked on MMAD-QA, GLLS offers a competitive and training-free alternative suitable for rapid deployment.
To further evaluate the robustness of GLLS, we extend the evaluation to MPDD, DTD, and DAGM (Table~\ref{tab:generalization_glls}). In the 1-shot setting, GLLS achieves an average accuracy of \textbf{91.01\%}, outperforming proprietary LMMs and several specialized CLIP-based visual anomaly detectors under this binary accuracy protocol~\cite{jeong2023winclip, zhou2024anomalyclip, gao2025adaptclip, deng2025abound}. This indicates that the proposed training-free pipeline leverages visual-logical cues that transfer across datasets.

\subsection{Ablation Studies and Component Analysis}
\label{sec:ablation}

We performed a systematic evaluation to isolate the contributions of individual components and to assess the trade-off between performance and computational cost.

\paragraph{Synergistic Contribution of Components}
Table~\ref{tab:ablation} presents the performance impact of removing key modules. The offline-only strategy, which uses static knowledge retrieval without active perception, often improves the base model but exhibits instability on the VisA dataset due to visual hallucinations. Conversely, the online-only configuration using dual-stream verification lacks the semantic grounding provided by the PVLA, and consequently struggles with defect classification. The full GLLS framework integrates both streams to achieve the best average results in most settings, indicating that rigorous offline standards benefit from active online verification.

Table~\ref{tab:stream_ablation} demonstrates the incremental benefits of the online dual-stream architecture. Adding the \textit{Fine-Grained \& Actions Stream} to the offline baseline yields consistent gains by acquiring high-resolution details, and the full GLLS achieves the highest performance. This result suggests that local evidence is more effective when contextualized by global reasoning.

Furthermore, Table~\ref{tab:pvla_ablation} isolates the contribution of the offline PVLA. Although PVLA generally outperforms uni-modal baselines, the performance drop observed with Qwen2.5-7B reveals a specific limitation: enforcing strict standards on passive, low-resolution inputs increases false positives. This finding indicates that rigorous offline knowledge depends on active online verification to successfully distinguish defects from noise.

\begin{table}[!t]
    \centering
    \caption{\textbf{Efficiency Analysis.} Latency breakdown. Total latency $T_{total} \approx \max(T_{glob}, T_{search}) + T_{verif}$ shows that the fine-grained search overhead is largely masked by the parallel logic checks. \textbf{G\&L}: Global \& Logic; \textbf{FG\&A}: Fine-Grained \& Actions.}
    \label{tab:efficiency_overall}
    
    \scriptsize
    \renewcommand{\arraystretch}{1.1}
    
    \begin{tabular}{llcccc}
        \toprule
        \multirow{2}{*}{\textbf{Model}} & \multirow{2}{*}{\textbf{Method}} & \multicolumn{2}{c}{\textbf{Dual Streams}} & \textbf{Reasoning} & \multirow{2}{*}{\textbf{Total}} \\
        \cmidrule(lr){3-4}
         &  & \textit{G\&L} & \textit{FG\&A} & \textit{Verification} &  \\
        \midrule
        \multirow{3}{*}{\shortstack{Qwen3\\4B}} 
         & Base & - & - & 0.74s & 0.74s \\
         & GLLS (0-shot) & 0.33s & 0.04s & 0.77s & 1.10s \\
         & GLLS (1-shot) & 0.31s & 0.07s & 0.90s & 1.21s \\
        \midrule
        \multirow{3}{*}{\shortstack{Qwen3\\8B}} 
         & Base & - & - & 0.77s & 0.77s \\
         & GLLS (0-shot) & 0.31s & 0.04s & 1.00s & 1.31s \\
         & GLLS (1-shot) & 0.33s & 0.08s & 1.35s & 1.68s \\
        \midrule
        \multirow{3}{*}{\shortstack{Qwen2.5\\7B}} 
         & Base & - & - & 0.85s & 0.85s \\
         & GLLS (0-shot) & 0.38s & 0.03s & 1.00s & 1.38s \\
         & GLLS (1-shot) & 0.32s & 0.08s & 1.52s & 1.84s \\
        \bottomrule
    \end{tabular}
\end{table}

\begin{table}[!t]
    \centering
    \caption{\textbf{Ablation Study on Fine-Grained \& Actions Stream.} Default setting ($K_{max}=3, N=50, C_p=2.0$) is highlighted in \colorbox{gray!15}{gray}. We report \textit{Total Time} for (b) and specific \textit{Search Time} for (c).}
    \label{tab:mcts_ablation_combined}
    
    \scriptsize
    \renewcommand{\arraystretch}{1.15} 
    
    \begin{minipage}[t]{0.48\linewidth}
        \centering
        \textbf{(a) Search Policy}\\
        \begin{tabular}{lccc}
            \toprule
            Strategy & MVTec & VisA & IoU \\
            \midrule
            Center Crop    & 84.2 & 75.1 & 65.2 \\
            Random         & 86.5 & 78.3 & 69.8 \\
            Sliding Window & 89.4 & 82.4 & 74.5 \\
            \textbf{GLLS (MCTS)} & \textbf{91.2} & \textbf{84.7} & \textbf{78.5} \\
            \bottomrule
        \end{tabular}
    \end{minipage}
    \hfill
    \begin{minipage}[t]{0.48\linewidth}
        \centering
        \textbf{(b) Evidence Budget ($K_{max}$)}\\
        \begin{tabular}{llcc}
            \toprule
            Budget & MVTec & VisA & \textbf{Total Time} \\
            \midrule
            1 Crop  & 88.5 & 81.2 & 1.15s \\
            \cellcolor{gray!15}\textbf{3 Crops} & \cellcolor{gray!15}\textbf{91.2} & \cellcolor{gray!15}\textbf{84.7} & \cellcolor{gray!15}\textbf{1.68s} \\
            5 Crops & 90.3 & 83.1 & 2.24s \\
            7 Crops & 90.1 & 82.0 & 2.89s \\
            \bottomrule
        \end{tabular}
    \end{minipage}

    \begin{minipage}[t]{0.48\linewidth}
        \centering
        \textbf{(c) Simulation Count ($N$)}\\
        \begin{tabular}{llcc}
            \toprule
            Count & MVTec & VisA & \textbf{Search Time} \\
            \midrule
            $N=10$  & 90.4 & 84.1 & 0.02s \\
            $N=30$  & 90.8 & 84.5 & 0.05s \\
            \cellcolor{gray!15}$\mathbf{N=50}$ & \cellcolor{gray!15}\textbf{91.2} & \cellcolor{gray!15}\textbf{84.7} & \cellcolor{gray!15}\textbf{0.08s} \\
            $N=100$ & 91.1 & 84.6 & 0.15s \\
            \bottomrule
        \end{tabular}
    \end{minipage}
    \hfill
    \begin{minipage}[t]{0.48\linewidth}
        \centering
        \textbf{(d) Exploration Const. ($C_p$)}\\
        \begin{tabular}{llcc}
            \toprule
            Value & MVTec & VisA & IoU \\
            \midrule
            $C_p=0.5$ & 90.5 & 84.0 & 77.6 \\
            $C_p=1.0$ & 90.9 & 84.3 & 78.1 \\
            \cellcolor{gray!15}$\mathbf{C_p=2.0}$ & \cellcolor{gray!15}\textbf{91.2} & \cellcolor{gray!15}\textbf{84.7} & \cellcolor{gray!15}\textbf{78.5} \\
            $C_p=4.0$ & 90.6 & 84.2 & 77.9 \\
            \bottomrule
        \end{tabular}
    \end{minipage}
\end{table}

\begin{figure}[tb]
  \centering
  \includegraphics[width=\linewidth]{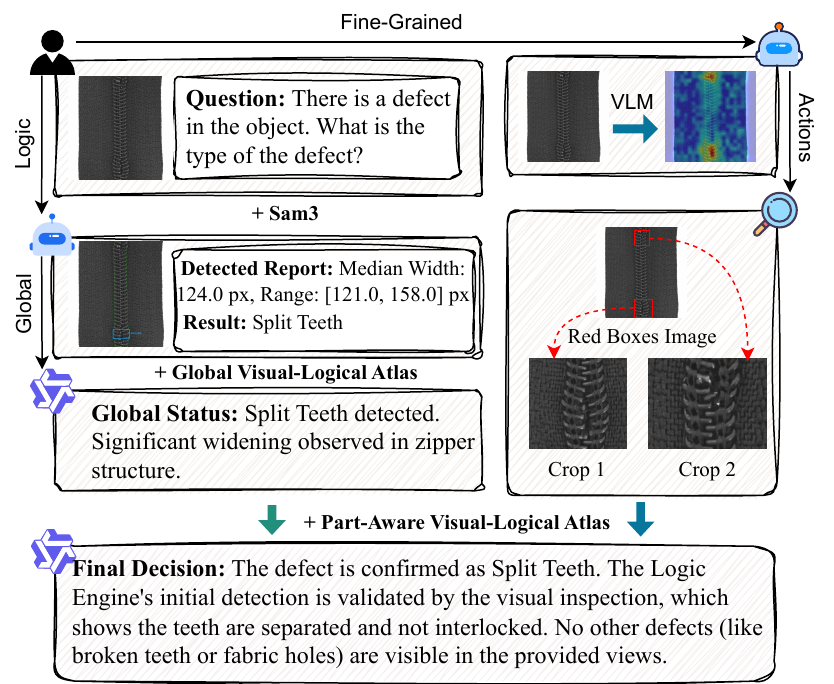}
    \caption{\textbf{Qualitative Visualization of the GLLS Inference Process.} 
Using a zipper inspection as an example, the framework illustrates the synthesis of global structural checks and local high-resolution evidence for verifiable diagnosis.}
  \label{fig:case}
\end{figure}
\paragraph{System Efficiency and Search Robustness}
Table~\ref{tab:efficiency_overall} reports the latency for the \textit{Global \& Logic} and \textit{Fine-Grained \& Actions} streams (denoted as \textit{Global Logic} and \textit{Local Search}, respectively). Results show that the latter operates faster than the former, so the additional search overhead remains small relative to final VLM verification. We further analyze the MCTS components in Table~\ref{tab:mcts_ablation_combined}: \textbf{(a) Search Policy:} MCTS-driven planning improves over sliding window and random baselines by leveraging heatmap heuristics to prioritize informative regions. \textbf{(b) Evidence Budget:} The VLM budget $K_{max}$ controls the cost-performance trade-off. We find $K_{max}=3$ to be a good operating point, as higher values increase latency with diminishing returns. \textbf{(c, d) Stability \& Efficiency:} The system shows limited sensitivity to these hyperparameters. Varying the simulation count or exploration constant results in marginal fluctuations. As shown in column (c), even with $N=100$, the search latency is only 0.15s, keeping the planning overhead small.

\subsection{Qualitative Visualization}
\label{sec:qualitative}
To demonstrate the interpretability of our framework, Figure~\ref{fig:case} visualizes a zipper inspection case involving a Split Teeth defect. In the Global \& Logic Stream, SAM 3 segments key components such as the zipper teeth, and the resulting visual primitives undergo deterministic PVLA checks. The logical report is combined with global multimodal knowledge and fed into the VLM to derive a preliminary conclusion on structural integrity. Concurrently, the Fine-Grained \& Actions Stream uses MCTS to capture high-resolution evidence crops revealing the separation. Final verification fuses global structural assessment with local visual details to produce an evidence-supported verdict traceable to visual observations.
\section{Conclusion}
\label{sec:concl}
We presented GLLS, a training-free dual-stream framework for verifiable industrial anomaly detection. The core idea is verification-first grounding, in which the system casts LMMs as evidence-guided auditors by providing part-level standards and making intermediate signals partially checkable through frozen external tools. By combining \textbf{Global \& Logic} for structural compliance with \textbf{Fine-Grained \& Actions} for detailed visual evidence, GLLS achieves strong performance on the MMAD-QA benchmark and shows cross-dataset generalization across several industrial domains including MPDD, DTD, and DAGM. Future work will focus on learning the search policy with verifiable rewards and expanding PVLA with richer contrastive edges to further improve discrimination accuracy.

\section*{Acknowledgement.}
\begin{sloppypar}
This work was supported by the National Natural Science Foundation of China (No.~62476124, 62302093), the Fundamental and Interdisciplinary Disciplines Breakthrough Plan of the Ministry of Education of China (JYB2025XDXM902), the Natural Science Foundation of Jiangsu Province (No.~BK20242015, BK20230833), the Gusu Innovation and Entrepreneur Leading Talents (No.~ZXL2025322), and Nanjing University-China Mobile Communications Group Co.,Ltd. Joint Institute.
\end{sloppypar}
\FloatBarrier
\bibliographystyle{splncs04}
\bibliography{main}

@misc{openai2025gpt5,
  title = {Introducing GPT-5},
  author = {{OpenAI}},
  year = {2025},
  month = {August 7},
  howpublished = {\url{https://openai.com/zh-Hans-CN/index/introducing-gpt-5/}},
  note = {Accessed: 2026-01-27}
}

@inproceedings{bigverdi2025perception,
  title={Perception tokens enhance visual reasoning in multimodal language models},
  author={Bigverdi, Mahtab and Luo, Zelun and Hsieh, Cheng-Yu and Shen, Ethan and Chen, Dongping and Shapiro, Linda G and Krishna, Ranjay},
  booktitle={Proceedings of the Computer Vision and Pattern Recognition Conference},
  pages={3836--3845},
  year={2025}
}

@inproceedings{wieler2007weakly,
  title={Weakly supervised learning for industrial optical inspection},
  author={Wieler, Matthias and Hahn, Tobias},
  booktitle={DAGM symposium in},
  volume={6},
  pages={11},
  year={2007}
}

@inproceedings{jezek2021deep,
  title={Deep learning-based defect detection of metal parts: evaluating current methods in complex conditions},
  author={Jezek, Stepan and Jonak, Martin and Burget, Radim and Dvorak, Pavel and Skotak, Milos},
  booktitle={2021 13th International congress on ultra modern telecommunications and control systems and workshops (ICUMT)},
  pages={66--71},
  year={2021},
  organization={IEEE}
}

@article{carion2025sam,
  title={Sam 3: Segment anything with concepts},
  author={Carion, Nicolas and Gustafson, Laura and Hu, Yuan-Ting and Debnath, Shoubhik and Hu, Ronghang and Suris, Didac and Ryali, Chaitanya and Alwala, Kalyan Vasudev and Khedr, Haitham and Huang, Andrew and others},
  journal={arXiv preprint arXiv:2511.16719},
  year={2025}
}

@inproceedings{baldassini2024makes,
  title={What makes multimodal in-context learning work?},
  author={Baldassini, Folco Bertini and Shukor, Mustafa and Cord, Matthieu and Soulier, Laure and Piwowarski, Benjamin},
  booktitle={Proceedings of the IEEE/CVF Conference on Computer Vision and Pattern Recognition},
  pages={1539--1550},
  year={2024}
}

@inproceedings{aota2023zero,
  title={Zero-shot versus many-shot: Unsupervised texture anomaly detection},
  author={Aota, Toshimichi and Tong, Lloyd Teh Tzer and Okatani, Takayuki},
  booktitle={Proceedings of the IEEE/CVF Winter Conference on Applications of Computer Vision},
  pages={5564--5572},
  year={2023}
}

@article{huang2024multimodal,
  title={Multimodal task vectors enable many-shot multimodal in-context learning},
  author={Huang, Brandon and Mitra, Chancharik and Arbelle, Assaf and Karlinsky, Leonid and Darrell, Trevor and Herzig, Roei},
  journal={Advances in Neural Information Processing Systems},
  volume={37},
  pages={22124--22153},
  year={2024}
}

@inproceedings{li2024improving,
  title={Improving Context Understanding in Multimodal Large Language Models via Multimodal Composition Learning.},
  author={Li, Wei and Fan, Hehe and Wong, Yongkang and Yang, Yi and Kankanhalli, Mohan S},
  booktitle={ICML},
  volume={3},
  number={6},
  pages={7},
  year={2024}
}

@article{bajcsy2018revisiting,
  title={Revisiting active perception},
  author={Bajcsy, Ruzena and Aloimonos, Yiannis and Tsotsos, John K},
  journal={Autonomous Robots},
  volume={42},
  number={2},
  pages={177--196},
  year={2018},
  publisher={Springer}
}

@article{browne2012survey,
  title={A survey of monte carlo tree search methods},
  author={Browne, Cameron B and Powley, Edward and Whitehouse, Daniel and Lucas, Simon M and Cowling, Peter I and Rohlfshagen, Philipp and Tavener, Stephen and Perez, Diego and Samothrakis, Spyridon and Colton, Simon},
  journal={IEEE Transactions on Computational Intelligence and AI in games},
  volume={4},
  number={1},
  pages={1--43},
  year={2012},
  publisher={IEEE}
}

@article{li2025iad,
  title={Iad-r1: Reinforcing consistent reasoning in industrial anomaly detection},
  author={Li, Yanhui and Cao, Yunkang and Liu, Chengliang and Xiong, Yuan and Dong, Xinghui and Huang, Chao},
  journal={arXiv preprint arXiv:2508.09178},
  year={2025}
}

@article{chao2025anomalyr1,
  title={Anomalyr1: A grpo-based end-to-end mllm for industrial anomaly detection},
  author={Chao, Yuhao and Liu, Jie and Tang, Jie and Wu, Gangshan},
  journal={arXiv preprint arXiv:2504.11914},
  year={2025}
}

@article{miao2025agentiad,
  title={AgentIAD: Tool-Augmented Single-Agent for Industrial Anomaly Detection},
  author={Miao, Junwen and Du, Penghui and Liu, Yi and Wang, Yu and Wang, Yan},
  journal={arXiv preprint arXiv:2512.13671},
  year={2025}
}

@inproceedings{li2025triad,
  title={Triad: Empowering LMM-based Anomaly Detection with Expert-guided Region-of-Interest Tokenizer and Manufacturing Process},
  author={Li, Yuanze and Yuan, Shihao and Wang, Haolin and Li, Qizhang and Liu, Ming and Xu, Chen and Shi, Guangming and Zuo, Wangmeng},
  booktitle={Proceedings of the IEEE/CVF International Conference on Computer Vision},
  pages={21917--21926},
  year={2025}
}

@inproceedings{zhang2025eiad,
  title={Eiad: Explainable industrial anomaly detection via multi-modal large language models},
  author={Zhang, Zongyun and Ruan, Jiacheng and Gao, Xian and Liu, Ting and Fu, Yuzhuo},
  booktitle={2025 IEEE International Conference on Multimedia and Expo (ICME)},
  pages={1--6},
  year={2025},
  organization={IEEE}
}

@article{zhao2025omniad,
  title={Omniad: Detect and understand industrial anomaly via multimodal reasoning},
  author={Zhao, Shifang and Lin, Yiheng and Han, Lu and Zhao, Yao and Wei, Yunchao},
  journal={arXiv preprint arXiv:2505.22039},
  year={2025}
}

@article{guan2025emit,
  title={Emit: Enhancing mllms for industrial anomaly detection via difficulty-aware grpo},
  author={Guan, Wei and Lan, Jun and Cao, Jian and Tan, Hao and Zhu, Huijia and Wang, Weiqiang},
  journal={arXiv preprint arXiv:2507.21619},
  year={2025}
}

@article{liao2025ad,
  title={AD-FM: Multimodal LLMs for Anomaly Detection via Multi-Stage Reasoning and Fine-Grained Reward Optimization},
  author={Liao, Jingyi and Su, Yongyi and Tu, Rong-Cheng and Jin, Zhao and Sun, Wenhao and Li, Yiting and Tao, Dacheng and Xu, Xun and Yang, Xulei},
  journal={arXiv preprint arXiv:2508.04175},
  year={2025}
}

@article{zeng2025lr,
  title={LR-IAD: Mask-Free Industrial Anomaly Detection with Logical Reasoning},
  author={Zeng, Peijian and Pang, Feiyan and Wang, Zhanbo and Yang, Aimin},
  journal={arXiv preprint arXiv:2504.19524},
  year={2025}
}

@inproceedings{chen2025can,
  title={Can multimodal large language models be guided to improve industrial anomaly detection?},
  author={Chen, Zhiling and Chen, Hanning and Imani, Mohsen and Imani, Farhad},
  booktitle={International Design Engineering Technical Conferences and Computers and Information in Engineering Conference},
  volume={89213},
  pages={V02BT02A051},
  year={2025},
  organization={American Society of Mechanical Engineers}
}

@article{gao2025adaptclip,
  title={Adaptclip: Adapting clip for universal visual anomaly detection},
  author={Gao, Bin-Bin and Zhou, Yue and Yan, Jiangtao and Cai, Yuezhi and Zhang, Weixi and Wang, Meng and Liu, Jun and Liu, Yong and Wang, Lei and Wang, Chengjie},
  journal={arXiv preprint arXiv:2505.09926},
  year={2025}
}

@article{deng2025abound,
  title={ABounD: Adversarial Boundary-Driven Few-Shot Learning for Multi-Class Anomaly Detection},
  author={Deng, Runzhi and Hu, Yundi and Zhang, Xinshuang and Wang, Zhao and Liu, Xixi and Dai, Wang-Zhou and Shan, Caifeng and Zhao, Fang},
  journal={arXiv preprint arXiv:2511.22436},
  year={2025}
}

@article{jiang2024mmad,
  title={Mmad: A comprehensive benchmark for multimodal large language models in industrial anomaly detection},
  author={Jiang, Xi and Li, Jian and Deng, Hanqiu and Liu, Yong and Gao, Bin-Bin and Zhou, Yifeng and Li, Jialin and Wang, Chengjie and Zheng, Feng},
  journal={arXiv preprint arXiv:2410.09453},
  year={2024}
}

@inproceedings{bergmann2019mvtec,
  title={MVTec AD--A comprehensive real-world dataset for unsupervised anomaly detection},
  author={Bergmann, Paul and Fauser, Michael and Sattlegger, David and Steger, Carsten},
  booktitle={Proceedings of the IEEE/CVF conference on computer vision and pattern recognition},
  pages={9592--9600},
  year={2019}
}

@article{hochstein2002view,
  title={View from the top: Hierarchies and reverse hierarchies in the visual system},
  author={Hochstein, Shaul and Ahissar, Merav},
  journal={Neuron},
  volume={36},
  number={5},
  pages={791--804},
  year={2002},
  publisher={Elsevier}
}

@article{ji2025autoiad,
  title={AutoIAD: Manager-Driven Multi-Agent Collaboration for Automated Industrial Anomaly Detection},
  author={Ji, Dongwei and Hu, Bingzhang and Zhou, Yi},
  journal={arXiv preprint arXiv:2508.05503},
  year={2025}
}

@inproceedings{zou2022spot,
  title={Spot-the-difference self-supervised pre-training for anomaly detection and segmentation},
  author={Zou, Yang and Jeong, Jongheon and Pemula, Latha and Zhang, Dongqing and Dabeer, Onkar},
  booktitle={European conference on computer vision},
  pages={392--408},
  year={2022},
  organization={Springer}
}

@article{bai2025qwen3,
  title={Qwen3-vl technical report},
  author={Bai, Shuai and Cai, Yuxuan and Chen, Ruizhe and Chen, Keqin and Chen, Xionghui and Cheng, Zesen and Deng, Lianghao and Ding, Wei and Gao, Chang and Ge, Chunjiang and others},
  journal={arXiv preprint arXiv:2511.21631},
  year={2025}
}

@article{hurst2024gpt,
  title={Gpt-4o system card},
  author={Hurst, Aaron and Lerer, Adam and Goucher, Adam P and Perelman, Adam and Ramesh, Aditya and Clark, Aidan and Ostrow, AJ and Welihinda, Akila and Hayes, Alan and Radford, Alec and others},
  journal={arXiv preprint arXiv:2410.21276},
  year={2024}
}

@article{yu2024visrag,
  title={Visrag: Vision-based retrieval-augmented generation on multi-modality documents},
  author={Yu, Shi and Tang, Chaoyue and Xu, Bokai and Cui, Junbo and Ran, Junhao and Yan, Yukun and Liu, Zhenghao and Wang, Shuo and Han, Xu and Liu, Zhiyuan and others},
  journal={arXiv preprint arXiv:2410.10594},
  year={2024}
}

@inproceedings{jeong2023winclip,
  title={Winclip: Zero-/few-shot anomaly classification and segmentation},
  author={Jeong, Jongheon and Zou, Yang and Kim, Taewan and Zhang, Dongqing and Ravichandran, Avinash and Dabeer, Onkar},
  booktitle={Proceedings of the IEEE/CVF conference on computer vision and pattern recognition},
  pages={19606--19616},
  year={2023}
}

@inproceedings{zhou2024anomalyclip,
  title={Anomalyclip: Object-agnostic prompt learning for zero-shot anomaly detection},
  author={Zhou, Qihang and Pang, Guansong and Tian, Yu and He, Shibo and Chen, Jiming},
  booktitle={International Conference on Learning Representations},
  volume={2024},
  pages={49705--49737},
  year={2024}
}

@article{bai2025qwen2,
  title={Qwen2.5-VL Technical Report},
  author={Bai, Shuai and Chen, Keqin and Liu, Xuejing and Wang, Jialin and Ge, Wenbin and Song, Sibo and Dang, Kai and Wang, Peng and Wang, Shijie and Tang, Jun and others},
  journal={arXiv preprint arXiv:2502.13923},
  year={2025}
}

@article{wang2025internvl3,
  title={Internvl3.5: Advancing open-source multimodal models in versatility, reasoning, and efficiency},
  author={Wang, Weiyun and Gao, Zhangwei and Gu, Lixin and Pu, Hengjun and Cui, Long and Wei, Xingguang and Liu, Zhaoyang and Jing, Linglin and Ye, Shenglong and Shao, Jie and others},
  journal={arXiv preprint arXiv:2508.18265},
  year={2025}
}

@article{yao2024minicpm,
  title={Minicpm-v: A gpt-4v level mllm on your phone},
  author={Yao, Yuan and Yu, Tianyu and Zhang, Ao and Wang, Chongyi and Cui, Junbo and Zhu, Hongji and Cai, Tianchi and Li, Haoyu and Zhao, Weilin and He, Zhihui and others},
  journal={arXiv preprint arXiv:2408.01800},
  year={2024}
}

@inproceedings{yu2026minicpm,
  title={Minicpm-v 4.5: Cooking efficient mllms via architecture, data, and training recipe},
  author={Yu, Tianyu and Wang, Zefan and Wang, Chongyi and Huang, Fuwei and Ma, Wenshuo and He, Zhihui and Cai, Tianchi and Chen, Weize and Huang, Yuxiang and Zhao, Ranchi and others},
  booktitle={Proceedings of the IEEE/CVF Conference on Computer Vision and Pattern Recognition},
  pages={11704--11715},
  year={2026}
}

@article{jaech2024openai,
  title={Openai o1 system card},
  author={Jaech, Aaron and Kalai, Adam and Lerer, Adam and Richardson, Adam and El-Kishky, Ahmed and Low, Aiden and Helyar, Alec and Madry, Aleksander and Beutel, Alex and Carney, Alex and others},
  journal={arXiv preprint arXiv:2412.16720},
  year={2024}
}

@article{guo2025deepseek,
  title={Deepseek-r1: Incentivizing reasoning capability in llms via reinforcement learning},
  author={Guo, Daya and Yang, Dejian and Zhang, Haowei and Song, Junxiao and Wang, Peiyi and Zhu, Qihao and Xu, Runxin and Zhang, Ruoyu and Ma, Shirong and Bi, Xiao and others},
  journal={arXiv preprint arXiv:2501.12948},
  year={2025}
}

@article{wei2022chain,
  title={Chain-of-thought prompting elicits reasoning in large language models},
  author={Wei, Jason and Wang, Xuezhi and Schuurmans, Dale and Bosma, Maarten and Xia, Fei and Chi, Ed and Le, Quoc V and Zhou, Denny and others},
  journal={Advances in neural information processing systems},
  volume={35},
  pages={24824--24837},
  year={2022}
}

@article{wang2025pixel,
  title={Pixel reasoner: Incentivizing pixel-space reasoning with curiosity-driven reinforcement learning},
  author={Wang, Haozhe and Su, Alex and Ren, Weiming and Lin, Fangzhen and Chen, Wenhu},
  journal={arXiv preprint arXiv:2505.15966},
  year={2025}
}

@article{zheng2025deepeyes,
  title={Deepeyes: Incentivizing" thinking with images" via reinforcement learning},
  author={Zheng, Ziwei and Yang, Michael and Hong, Jack and Zhao, Chenxiao and Xu, Guohai and Yang, Le and Shen, Chao and Yu, Xing},
  journal={arXiv preprint arXiv:2505.14362},
  year={2025}
}

@misc{song2025injectingdomainspecificknowledgelarge,
      title={Injecting Domain-Specific Knowledge into Large Language Models: A Comprehensive Survey}, 
      author={Zirui Song and Bin Yan and Yuhan Liu and Miao Fang and Mingzhe Li and Rui Yan and Xiuying Chen},
      year={2025},
      eprint={2502.10708},
      archivePrefix={arXiv},
      primaryClass={cs.CL},
      url={https://arxiv.org/abs/2502.10708}, 
}

@article{mecklenburg2024injecting,
  title={Injecting new knowledge into large language models via supervised fine-tuning},
  author={Mecklenburg, Nick and Lin, Yiyou and Li, Xiaoxiao and Holstein, Daniel and Nunes, Leonardo and Malvar, Sara and Silva, Bruno and Chandra, Ranveer and Aski, Vijay and Yannam, Pavan Kumar Reddy and others},
  journal={arXiv preprint arXiv:2404.00213},
  year={2024}
}

@inproceedings{jin2025logicad,
  title={Logicad: Explainable anomaly detection via vlm-based text feature extraction},
  author={Jin, Er and Feng, Qihui and Mou, Yongli and Lakemeyer, Gerhard and Decker, Stefan and Simons, Oliver and Stegmaier, Johannes},
  booktitle={Proceedings of the AAAI Conference on Artificial Intelligence},
  volume={39},
  number={4},
  pages={4129--4137},
  year={2025}
}

@article{li2025lad,
  title={Lad-reasoner: Tiny multimodal models are good reasoners for logical anomaly detection},
  author={Li, Weijia and Chu, Guanglei and Chen, Jiong and Xie, Guo-Sen and Shan, Caifeng and Zhao, Fang},
  journal={arXiv preprint arXiv:2504.12749},
  year={2025}
}
\end{document}